\documentclass[runningheads]{llncs}

\usepackage{eccv}

\usepackage{eccvabbrv}

\usepackage{graphicx}
\usepackage{booktabs}
\usepackage{amsfonts}
\usepackage{adjustbox}
\usepackage{multirow}
\usepackage{amsmath}
\usepackage[accsupp]{axessibility}  

\usepackage{hyperref}

\usepackage{orcidlink}

\begin{document}

\title{ClickSeg3D: Few-Click Interactive Segmentation via Semantic Embeddings} 

\titlerunning{One Click Model}

\author{
Xueyang Kang\inst{1}\orcidlink{0000-0001-7159-676X}\thanks{Equal contribution.}
\and
Zijian Yu\inst{2}\orcidlink{0009-0005-0253-4086}$^{\star}$
\and
Kourosh Khoshelham\inst{1}\orcidlink{0000-0001-6639-1727}
\and
Liangliang Nan\inst{3}\orcidlink{0000-0002-5629-9975}
}

\authorrunning{X. Kang et al.}

\institute{
School of Electrical and Electronic Engineering, Nanyang Technological University
\and
University of Science and Technology of China
\and
Faculty of Architecture and the Built Environment, Delft University of Technology
\email{
xueyang.kang@ntu.edu.sg,
zijian.yu@mail.ustc.edu.cn,
}
}

\maketitle

\begin{abstract}
Interactive segmentation allows efficient label generation by leveraging user-provided clicks to progressively refine predictions, which is critical when fully supervised labels are costly or generalization to unseen classes is needed. Existing 3D interactive methods are limited: most operate sequentially, predicting only one object per iteration with binary masks, while several recent approaches depend on 2D foundation models and camera alignment to bridge the 2D-3D gap. To address these limitations, we propose a novel interactive segmentation framework that operates directly on sparse, randomly downsampled 3D points and processes multiple object clicks in a single forward pass. Our framework consists of a point Transformer–based encoder and a hierarchical mask decoder, which integrates multi-level crop-and-merge operations conditioned on learnable semantic embeddings. Unlike prior interactive approaches that require repeated model updates after each manually corrective click, our method jointly reasons over all click queries, modeling inter-instance relationships and refining both spatial masks and semantic predictions through spatial and semantic embeddings. Extensive experiments demonstrate that our model improves the mIoU metric by over 20\% compared to strong baselines and achieves 8–10\% gains under cross-dataset evaluation for a one-click per instance setting, often requiring only a single click per object. Our approach provides a generalizable and efficient solution for interactive 3D instance segmentation, particularly suitable for real-time applications such as robotic manipulation, navigation, and rapid 3D semantic annotation. Public code is available here: \href{https://github.com/alexandor91/OneClickSeg.git}{Project Repository}
\end{abstract}

\section{Introduction}
Interactive 2D segmentation has advanced significantly with models like the Segment Anything Model (SAM) \cite{kirillov2023segment}. However, extending such methods to 3D segmentation in open-world, zero-shot, or few-shot settings remains underexplored. Current 3D approaches often leverage 2D foundation models to guide 3D segmentation, as in SAM3D \cite{yang2023sam3d} and SAMPro3D \cite{xu2023sampro3d}, using prompts such as text \cite{peng2023openscene, zhu2024open, qinLangSplat3DLanguage2024}, multi-view 2D masks \cite{guo2024sam2point}, or 3D clicks \cite{choi2025click}. Recently, Meta has reconstructed 3D object geometry and texture from a single image, handling occlusion and clutter via large-scale visually grounded data and a synthetic-to-real training pipeline, as ``SAM 3D'' \cite{sam3dteam2025sam3d3dfyimages}. Despite these efforts, generalization to complex, unannotated 3D scenes is limited, motivating more scalable and efficient solutions.

State-of-the-art methods face key challenges in efficiency and multi-object segmentation. Interobject3D \cite{lang2024iseg} processes each user click independently, resulting in slow inference for multiple objects. Agile3D \cite{yue2023agile3d} improves efficiency via a lightweight decoder and inter-click relationships but still relies on iterative refinements with human correcting feedback, which is computationally expensive and limits scalability, especially for unseen datasets. Point-SAM \cite{zhou2024point} generates pseudo-2D masks for 3D point clouds, partially addressing the scarcity of labeled data. Other works lift 2D SAM masks to 3D via direct adaptation \cite{yang2023sam3d}, implicit radiance fields \cite{goel2023interactive}, or explicit Gaussian representations \cite{choi2025click}, yet they depend on accurate camera pose alignments and are sensitive to sparse viewpoints and occlusion, causing inconsistencies between 2D masks and 3D labels.

To overcome these limitations, we propose a novel 3D interactive segmentation framework trained directly on sparse, randomly downsampled points to simulate user clicks. Our method segments multiple objects in a single forward pass, avoiding iterative refinement. We employ a Point Transformer v3 \cite{wu2024ptv3} encoder to extract rich features from both the scene and click points, with a shared point encoder. The segmentation decoder incorporates a multi-level crop-and-merge mechanism, producing accurate instance masks via a mask head and class head. Learnable semantic prototype embeddings condition query features with global context, enhancing attention, robustness to click variations, and generalization to unseen datasets.

\begin{itemize}
\item \textbf{Decoupled Single-Pass Architecture.} 
We propose a unified framework that explicitly disentangles point feature extraction from multi-stage mask refinement. Unlike prior iterative pipelines, our design performs efficient and fine-grained multi-object segmentation in a single forward pass, reducing computational overhead while preserving segmentation quality.

\item \textbf{Sparse Interaction-Driven Training Strategy.} 
We introduce a training paradigm based on sparsely and randomly downsampled points to simulate realistic user interactions. This strategy improves robustness to varying click numbers and spatial distributions, ensuring stable performance across diverse interactive settings while maintaining training efficiency.

\item \textbf{Multi-Stage Crop-and-Merge Decoder with Semantic Prototype Embeddings.} 
We design a progressive crop-and-merge decoder that refines instance masks from coarse to fine granularity. By leveraging learnable semantic prototype embeddings to condition and update query features, the decoder enhances mask precision without increasing computational cost, leading to improved segmentation accuracy and stronger generalization to unseen categories.
\end{itemize}
\section{Related  Work}
\label{sec:related-work}
Traditional 3D segmentation models rely heavily on dense annotations and often fail to generalize to unseen data, limiting their practicality for the challenging task of \emph{open-world querying} \cite{peng2023openscene}, which requires class-agnostic reasoning. Such generalization is especially critical for downstream embodied applications, where robotic agents must perceive, navigate, and manipulate within diverse and previously unseen environments under limited supervision \cite{wong2025survey}. Earlier efforts in object-level semantic mapping have likewise emphasized the importance of robust object-centric reasoning and consistent association across views, which directly motivates class-agnostic, instance-aware 3D understanding \cite{kang2019robust}. To overcome these challenges, recent works leverage prior knowledge from 2D foundation models, fusing multi-view 2D masks into 3D representations in few-shot or self-supervised settings \cite{Huang2023Segment3D, yinSAI3DSegmentAny2024, peng2023sam}. In parallel, other methods integrate iterative human feedback into the 3D segmentation pipeline, employing cross-attention between click-based queries and scene features to refine masks. Nonetheless, their dependence on precise prompts and repeated interactions underscores the need for more efficient click-based strategies that can reliably separate adjacent objects from the background, especially in ambiguous cases.

\noindent\textbf{3D segmentation with 2D masks.} Leveraging 2D foundation models has become a common strategy in 3D segmentation \cite{takmazOpenMask3DOpenVocabulary3D2023}, where 2D multi-view segmentation results are lifted into 3D space. Recent techniques \cite{choi2025click,yan2024maskclustering,nguyenOpen3DISOpenVocabulary3D2024,liu2024sanerf,engelmannOpenNeRFOpenSet2024,huangGaussianFormerSceneGaussians2024,lan2DGuided3DGaussian2023,guo2023sam,shi2023language} exploit models such as the Segment Anything Model (SAM) \cite{kirillovSegmentAnything2023} to generate 2D masks from image sequences, which then supervise the learning of 3D representations and semantic labels. For example, SAM3D \cite{yang2023sam3d} and SAMPro3D \cite{xu2025sampro3d} jointly combine 2D SAM masks with 3D superpoints to produce fine-grained point-level segmentation. Similarly, Segment3D \cite{Huang2023Segment3D} trains a 3D segmentation model on projected 2D SAM masks and partially labeled 3D points, using pseudo-labeling to enable zero-shot and open-set segmentation from text prompts. While these methods reduce the need for dense 3D annotations, they still require large numbers of posed multi-view images per scene, making the overall process cumbersome. To mitigate the burden of dense view acquisition, recent advances in geometry-aware view synthesis and panorama-then-video scene generation \cite{kang2025multi,kang2025look} explore ways to hallucinate additional views from sparse observations, complementing classical pipelines that reconstruct coarse 3D geometry directly from low-cost 2D sensors \cite{kang20183d}.

\noindent\textbf{3D segmentation with language.} The advent of large vision-language models has enabled direct labeling of 3D points via language prompts, using either CLIP-distilled 3D feature fields \cite{radford2021learning} or large language models such as \cite{jatavallabhulaConceptFusionOpensetMultimodal2023,qinLangSplat3DLanguage2024,boudjoghra20243d,wu2024opengaussian,shi2024language,zheng2024gaussiangrasper}. For instance, OVIR-3D \cite{luOVIR3DOpenVocabulary3D2023} lifts 2D region proposals into 3D to support language-driven scene queries. Similarly, LERF \cite{lerf2023} and DFF \cite{kobayashi2022distilledfeaturefields} integrate language models with NeRF by optimizing an additional CLIP-aligned feature field, while OpenNeRF \cite{engelmann2024opennerf} replaces CLIP with pixel-aligned features from OpenSeg \cite{ghiasiScalingOpenVocabularyImage2022} to achieve fine-grained view segmentation. Although these approaches demonstrate strong results, their reliance on 3D representation priors limits generalization across diverse scenes and typically demands dense multi-view coverage, leading to significant training complexity.

\begin{figure*}[!ht]
    \centering
    \includegraphics[width=\linewidth]{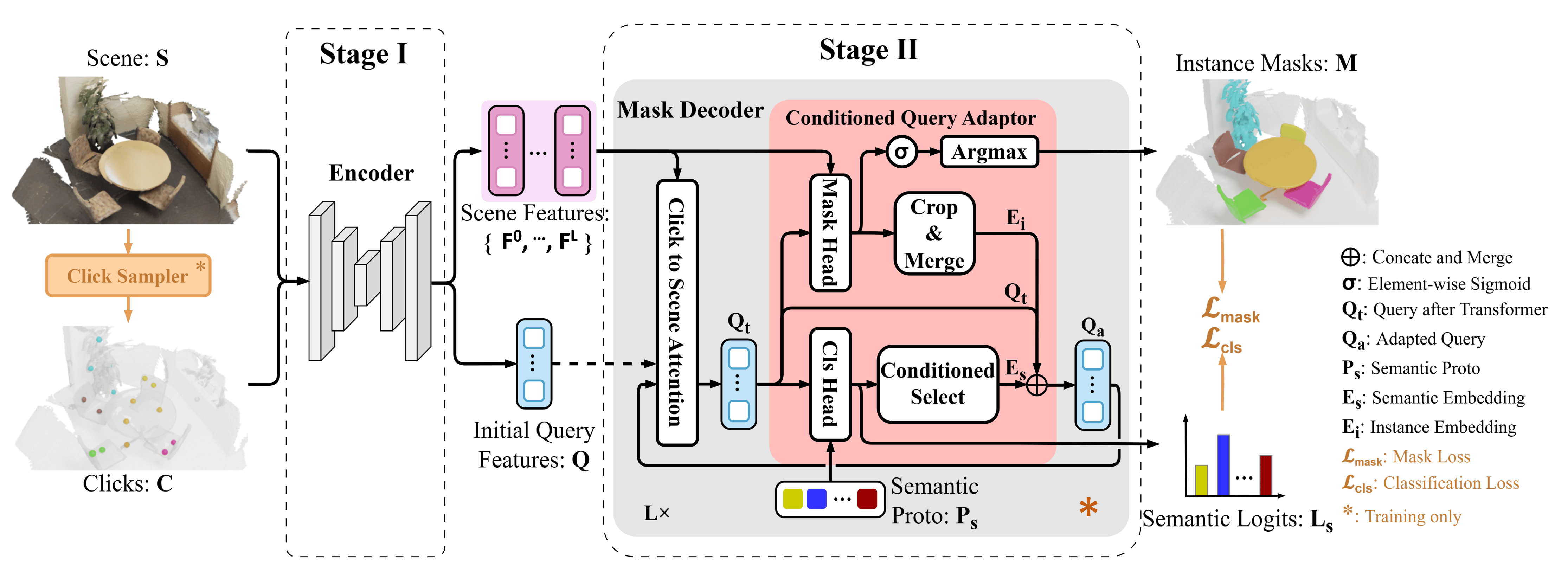}
    \caption{
        Overview of our click-based instance segmentation framework. Given a scene $\mathbf{S}$ with user-provided clicks $\mathbf{C}$, the \textbf{scene encoder} extracts multi-scale scene features $\mathbf{\{F_0, ..., F_L\}}$, while the query encoder produces query features $\mathbf{Q}$. 
        The transformer block refines these features into $\mathbf{Q_t}$, which the \textit{Conditioned Query Adaptor} further refines into $\mathbf{Q_s}$ using the semantic prototype $\mathbf{P_s}$ and semantic embedding $\mathbf{E_s}$. 
        The mask decoder employs a mask head and classification head to generate instance embeddings $\mathbf{E_i}$, producing instance masks $\mathbf{M}$. 
        The segmentation is optimized with mask loss $\mathcal{L}_{\text{mask}}$ and classification loss $\mathcal{L}_{cls}$. The dashed line indicates the initial query feature values at the first time step.}
    \label{fig:overview}
\end{figure*}

\noindent\textbf{3D segmentation with sparse 3D prompts.} Compared to other approaches for fine-grained point cloud labeling, 3D segmentation with sparse 3D prompts (i.e., interactive 3D segmentation) offers a more efficient interaction paradigm by using bounding boxes or seed points as inputs. Some methods expand labels iteratively from a few input seed points \cite{jain2024gaussiancut,osepBetterCallSAL2024,zhangRefiningSegmentationFly2024,choiIDet3DEfficientInteractive2023}, while others adopt a hierarchical strategy, progressively refining labels from coarse to fine \cite{zhou2024point,liu2023clickseg,lang2024iseg,kang2026hierarchical}. InterObject3D \cite{kontogianni2022interObj3d} extends Minkowski UNet by adding a channel for projected point prompts as a binary mask, but it is restricted to a single instance per inference and requires re-running the full model for each refinement, making it inefficient. AGILE3D \cite{yue2023agile3d}, inspired by Mask3D \cite{schultMask3DMaskTransformer2023}, generalizes interactive prompts to multiple objects, requiring only the decoder stage for refinement. Alternatively, SAM2Point \cite{guo2024sam2point} converts voxels around clicked points into 2D slices, applies SAM2 for segmentation, and then unprojects the labels back to 3D. Beyond these prompt-driven schemes, robust geometric descriptors, such as equivariant SE(3) features learned from 2D surfel-based representations for point cloud registration \cite{kang20252d}, offer a complementary avenue for stabilizing sparse 3D prompts under viewpoint variation. While these methods enable more convenient interaction, their segmentation quality degrades with sparse or noisy inputs, and results remain unstable due to high sensitivity to the choice of initial prompt points.
\section{Method}
Our model consists of a pre-trained PTv3 \cite{wu2024ptv3} backbone and a learnable decoder that jointly processes all click-per-instance inputs at once. \cref{fig:overview} illustrates our two-stage pipeline for interactive 3D segmentation based on clicks. Given a scene and user-provided clicks, the framework extracts multi-scale scene and query point features through pre-trained encoders. A conditional query adaptor then refines the queries with semantic and instance spatial cues, enabling the mask decoder to progressively transform coarse predictions into fine-grained instance masks in a single forward pass. The final output is an interactive segmentation result produced with lightweight post-processing.

\subsection{Interactive 3D Segmentation Preliminaries}
Given a 3D scene $\mathbf{S}$, interactive segmentation aims to iteratively refine instance masks $\mathbf{M}$ with additional click guidance $\mathbf{C}$. State-of-the-art training pipelines \cite{kontogianni2022interObj3d,yue2023agile3d} rely on iterative inference, where each step uses the previous segmentation as input and adds corrective clicks in the largest error regions, either positive or negative clicks, to refine the results. This process significantly increases training time, often taking several times longer than training for direct instance segmentation. To eliminate repeated forward passes, we reformulate interactive 3D segmentation as a one-click-per-instance multi-object prediction problem. Under this formulation, the positive click of one instance implicitly serves as negative supervision for other instances. We adapt the click sampler and employ a single forward-pass decoder for coarse segmentation, as illustrated by the mask decoder in Stage II of \cref{fig:overview}.

\setlength{\parindent}{0pt}
\textbf{Sampling for Training.} Given a 3D scene \( \mathbf{S} \in \mathbb{R}^{N \times 3} \) with \( N \) points, we use Farthest Point Sampling (FPS) to generate simulated click points \( \mathbf{C} \) for both one-click-to-one-instance and multi-click-to-one-instance segmentation tasks. To ensure spatially even sampling, FPS with augmented transformations is applied, selecting the farthest point from the sampled set to achieve uniform coverage of the instance geometry. On average, 1–15 points per instance are sampled as click candidates for training, balancing diversity with efficiency. During training, a random subset of sampled points per instance is selected as input click $\mathbf{C}$, simulating user interaction. This enhances robustness by making the model invariant both to the number and spatial distribution of clicks during inference.


\textbf{Post-Processing of Segmentation Results.} During inference, the segmentation mask result $\mathbf{M} \in \mathbb{R}^{N \times K}$ assigns each of the $N$ points in the input scene to one of $K$ classes, conditioned on the input click $\mathbf{C}$. Points corresponding to the same click are labeled under the same instance class. Applying $ArgMax(\cdot)$ to mask $\mathbf{M}$ ensures that masks of different clicks are mutually exclusive. These masks are then merged according to the user-defined instance-per-click relation to form the final interactive segmentation. When the user provides multiple clicks ${\{\mathbf{C}_{i}\}, i \in K}$, for a single instance, all clicks $\{\mathbf{C}_{i}\}$ are grouped into one cluster. The model treats these $K$ clicks as intra-instance guidance, while preserving inter-instance semantic separation across different clicks.

\subsection{Click-Guided Point Cloud Instance Segmentation}

Our model comprises two main components: a point Transformer encoder, shared by both scene points and query clicks, that extracts per-point features and transforms them into semantic query features, and a mask decoder that hierarchically refines these queries to generate instance masks.

\textbf{Scene Encoder.} Our scene encoder $\psi$ is a Point Transformer–based encoder. Given a 3D scene $\mathbf{S} \in \mathbb{R}^{N \times 3}$ with $N$ points as input, it extracts multiscale per-point features $\{\mathbf{F}^0, \ldots, \mathbf{F}^L\}$ from $0$ to $L$ levels.
\begin{equation}
    \mathbf{\{F^0,...,F^L\}} = \psi(\mathbf{S}),
    \label{eq:encoder-scene}
\end{equation}
where $\{\mathbf{F}_j^i \in \mathbb{R}^{D_{i}} \mid i \in [1, \ldots, L], \, j \in [1, \ldots, N]\}$, $L$ is the number of scales, $D_{i}$ is the feature dimension at scale $i$, and $N$ is the number of points.

\textbf{Query Encoder.} The shared encoder $\psi$ also converts each user click $\mathbf{C}$ into a query feature $\mathbf{Q}$. For each click, the nearest voxel in $\mathbf{F}^L$ at $L$th resolution is located using a KNN lookup, and its feature is assigned as $\mathbf{Q}$:
\begin{equation}
    \mathbf{Q} = \mathbf{F}^L\left[\textit{Lookup}(\mathbf{C},\mathbf{S})\right],
    \label{eq:index}
\end{equation}
where $\mathbf{C} \in \mathbb{R}^{K \times 3}$ and $\mathbf{Q} \in \mathbb{R}^{K \times D_q}$, with $K$ denoting the number of clicks and $D_q$ the query feature dimension.
After initialization, each query feature is iteratively refined with per-point features through the decoder’s attention.

\textbf{Mask Decoder.} Our mask decoder contains $L$ stages, each consisting of a transformer block $\mathcal{T}$ and a conditioned query adaptor $\mathcal{G}$. In stage $i \in \{1, \ldots, L\}$, the transformer block takes as input the scene feature $\mathbf{F}^i$ and the adapted query feature $\mathbf{Q}_a^{i-1}$ from the previous stage. First, cross-attention between $\mathbf{Q}_a^{i-1}$ and $\mathbf{F}^i$ is computed. The resulting query is then processed by a self-attention layer and a feed-forward network to produce the transformer output $\mathbf{Q}_t^i$. Finally, $\mathbf{Q}_t^i$ is passed through the conditioned query adaptor, detailed as,
\begin{align}
    \mathbf{Q}_t^i = \mathcal{T}(\mathbf{Q}_a^{i-1},\mathbf{F}^i), \\
    \mathbf{Q}_a^i = \mathcal{G}(\mathbf{Q}_t^i, \mathbf{F}^L),
\label{eq:q_feature}
\end{align}
where the initial query feature $\mathbf{Q}_a^{0}$ is set to be the output of the query encoder $\mathbf{Q}$. This design aims to enable interaction between the click queries themselves and between queries and the point features. As shown in \cref{fig:overview}, each decoder layer consists of a click-to-scene attention module (C2S), a click-to-click attention module (C2C), a feed-forward network (FFN), and a mask module (MM). First, we compute Fourier positional encoding \cite{li2021learnable} for voxel positions as $...$ to model the spatial information of scene points. The C2S performs cross-attention from click queries to point features, which enables click queries to extract information from relevant regions in the point cloud. In the C2C, we let each click query self-attend to each other to realize inter-query communications. The C2C is followed by an FFN. Finally, the module MM uses queries for classification and segmentation, and refines the query using the semantic prototype $\mathrm{P_s}$ and coarse segmentation results of the initial output. 

\subsection{Conditional Query Adaptor.}
\label{sec:method-CQA}

As shown in Fig.~\ref{fig:overview}, the conditioned query adaptor is designed to refine query features and compute segmentation results. Prior work refines masks by applying cross-attention to multi-scale scene features. However, this multi-scale approach loses information through the implicit feature mapping layer, since queries are transformed from dense implicit vectors into sparse explicit click coordinates. To address this, we introduce the conditioned query adaptor, which refines sparse click features using both explicit learnable semantic and spatial information. The mask decoder contains multiple stages; at each stage $i$, the adaptor takes as input the query feature from the last transformer block $\mathbf{Q}_t^i$ and the $L$th-level scene feature $\mathbf{F}^L$. First, segmentation outputs are predicted using a mask head and a class head, where classification relies on learnable semantic prototype embeddings. Next, a spatial embedding $\mathbf{E}_p^i$ and a semantic embedding $\mathbf{E}_s^i$ are learned for each instance. Finally, $\mathbf{E}_p^i$, $\mathbf{E}_s^i$, and $\mathbf{Q}_t^i$ are concatenated and fused with a lightweight MLP to produce the adapted query feature $\mathbf{Q}_a^i$.

\textbf{Mask Head.} Given the scene feature $\mathbf{F}^L$ and the query feature $\mathbf{Q}_t$, the segmentation mask is defined as,
\begin{equation}
    \mathbf{M}^i = \mathbf{F}^L \otimes \phi_m(\mathbf{Q}_t^i)^T,
    \label{eq:mask}
\end{equation}
where $\mathbf{M}^i\in\mathbb{R}^{N\times K}$, $N$ is the number of scene points, $\phi_m$ denotes the MLP layer and $\otimes$ is matrix multiplication.

\textbf{Class Head.} Inspired by open-vocabulary segmentation \cite{luOVIR3DOpenVocabulary3D2023}, we parameterize the segmentation kernel as a set of learnable semantic prototype embedding $\mathbf{P}_s$. For query feature $\mathbf{Q}_t^i$, the classification result is computed as,
\begin{equation}
    \mathbf{Z}^i = \mathbf{P}_s \otimes \phi_c(\mathbf{Q}_t^i)^T,
    \label{eq:class}
\end{equation}
where $\mathbf{Z}^i \in \mathbb{R}^{N_c \times K}$, with $N_c$ denoting the number of classes and $\phi_c$ representing the MLP layer.

\textbf{Query Adaptation.} After obtaining the segmentation result in stage $i$, we use it to refine the information from the coarse stage $i$ to the fine stage $i+1$. At each stage, we reinitialize a spatial embedding $\mathbf{E}_p^i \in \mathbb{R}^{256}$ and a semantic embedding $\mathbf{E}_s^i \in \mathbb{R}^{d_h}$ for each instance. The spatial embedding $\mathbf{E}_p^i$ is computed by cropping and aggregating the scene feature $\mathbf{F}_i \in \mathbb{R}^{256}$ using mask pooling with the segmentation mask $\mathbf{M}^i$. The semantic embedding $\mathbf{E}_s^i$ is selected from semantic prototype embeddings $\mathbf{P}_s \in \mathbb{R}^{d_h}$ conditioned on the class prediction $\mathbf{Z}^i$. Finally, we calculate the query feature $\mathbf{Q}_t^i \in \mathbb{R}^{256}$ with both spatial and semantic embeddings $\mathbf{E}_p^i$ and $\mathbf{E}_s^i$, then obtain the adapted query feature $\mathbf{Q}_a^i \in \mathbb{R}^{256}$ using a small MLP:

\begin{equation}
    \mathbf{Q}_a^i = \phi_q(\mathrm{Concat}(\mathbf{Q}_t^i,\mathbf{E}_p^i,\mathbf{E}_s^i)),
    \label{eq:query-adaptor}
\end{equation}
where $\phi_q$ is an MLP that maps the concatenated embedding to the query embedding.

\subsection{Loss Function.} We supervise the network with a binary cross-entropy $\mathcal{L}_\mathrm{BCE}$, Rolling the Dice loss \cite{elmqvist2008rolling} $\mathcal{L}_\mathrm{Dice}$, and a classification loss, while the classification loss is standard cross-entropy $\mathcal{L}_\mathrm{CE}$:
\begin{equation}
    \mathcal{L}_{total} = \lambda_\mathrm{BCE}\mathcal{L}_\mathrm{BCE} + \lambda_\mathrm{Dice}\mathcal{L}_\mathrm{Dice} +  \lambda_\mathrm{CE}\mathcal{L}_\mathrm{CE},
    \label{eq:loss}
\end{equation}
where $\lambda_\mathrm{BCE}$, $\lambda_\mathrm{Dice}$ and $ \lambda_\mathrm{CE}$ are the balancing loss weights.

The \textbf{binary cross-entropy loss} $\mathcal{L}_\mathrm{BCE}$ supervises point-wise mask prediction by penalizing discrepancies between the predicted mask probability $\hat{m}_i \in [0,1]$ and the ground-truth mask label $m_i \in \{0,1\}$. It encourages accurate foreground–background classification at the point level and is defined as
\begin{equation}
\mathcal{L}_\mathrm{BCE} = -\frac{1}{N}\sum_{i=1}^{N} \big[ m_i \log(\hat{m}_i) + (1-m_i)\log(1-\hat{m}_i) \big],
\label{eq:bce}
\end{equation}
where $N$ denotes the number of points.

The \textbf{Dice loss} $\mathcal{L}_\mathrm{Dice}$ measures the overlap between the predicted mask $\hat{m}$ and the ground-truth mask $m$, alleviating class imbalance by directly optimizing region-level consistency. Following the Dice formulation~\cite{elmqvist2008rolling}, it is expressed as
\begin{equation}
\mathcal{L}_\mathrm{Dice} = 1 - \frac{2 \sum_{i=1}^{N} \hat{m}_i m_i + \epsilon}{\sum_{i=1}^{N} \hat{m}_i + \sum_{i=1}^{N} m_i + \epsilon},
\label{eq:dice}
\end{equation}
where $\epsilon$ is a small constant for numerical stability.

The \textbf{classification loss} $\mathcal{L}_\mathrm{CE}$ is a standard cross-entropy loss applied to object-level or instance-level class predictions. It enforces semantic consistency by comparing the predicted class distribution $\hat{\mathbf{y}}$ with the ground-truth label $\mathbf{y}$:
\begin{equation}
\mathcal{L}_\mathrm{CE} = -\sum_{c=1}^{C} y_c \log(\hat{y}_c),
\label{eq:ce}
\end{equation}
where $C$ is the number of semantic classes.

\section{Experiments \& Results}
\noindent \textbf{Datasets.} For evaluation, we use the indoor dataset ScanNet40~\cite{dai2017scannet}, which contains 40 main indoor classes, and KITTI360~\cite{Liao2022PAMI} for outdoor evaluation. Furthermore, we perform cross-dataset evaluation, training on ScanNet40 \cite{dai2017scannet} with 40 classes and testing on S3DIS \cite{hou3DSIS3DSemantic2019} and KITTI360 \cite{Liao2022PAMI}, and The cross-evaluation results, measured by IoU as a function of click number and the average number of clicks required to reach a given IoU threshold, are reported in \cref{tab:multi_object_cross}.

\noindent \textbf{Evaluation Metrics.}
We evaluate 3D instance segmentation performance using mean Intersection-over-Union (mIoU) and mean Average Precision (mAP). The mIoU metric measures the overlap between predicted and ground-truth points at the semantic level, averaged across all classes, and is widely used for assessing segmentation quality. In addition, we report mIoU@N, where N denotes the number of clicks. In addition, we also report mAP, which evaluates detection and segmentation jointly by computing precision–recall curves under an Intersection-over-Union (IoU) matching criterion. The mAP is commonly reported at multiple mIoU overlap thresholds (e.g., mAP@0.25 and mAP@0.5).

\noindent \textbf{Implementation Details.}
We first designed an offline training sampler to simulate user-guided clicks, which can be pre-cached before training to eliminate user intervention during the training phase. For scene training, we randomly sample around 30--50 points to mimic user clicks. The embedding dimensions for both spatial and semantic embeddings are set to 256, while the number of semantic prototype embeddings is 35. The decoder is composed of four repeated stages.  

For baselines, we primarily adopt click-prompt-based methods, including InterObj3D~\cite{kontogianni2022interObj3d} and its improved variant InterObj3D++ as described in Agile3D~\cite{yue2023agile3d}. We further include Agile3D~\cite{yue2023agile3d} itself as a baseline, along with Point-SAM~\cite{zhou2024point}, which takes point clouds as input, and SAM2Point~\cite{guo2024sam2point}, which uses 3D point prompts as input.

\subsection{Baseline Comparison}
\begin{figure*}[!th]
    \centering
\vspace*{0.12cm} 
\includegraphics[width=\linewidth]{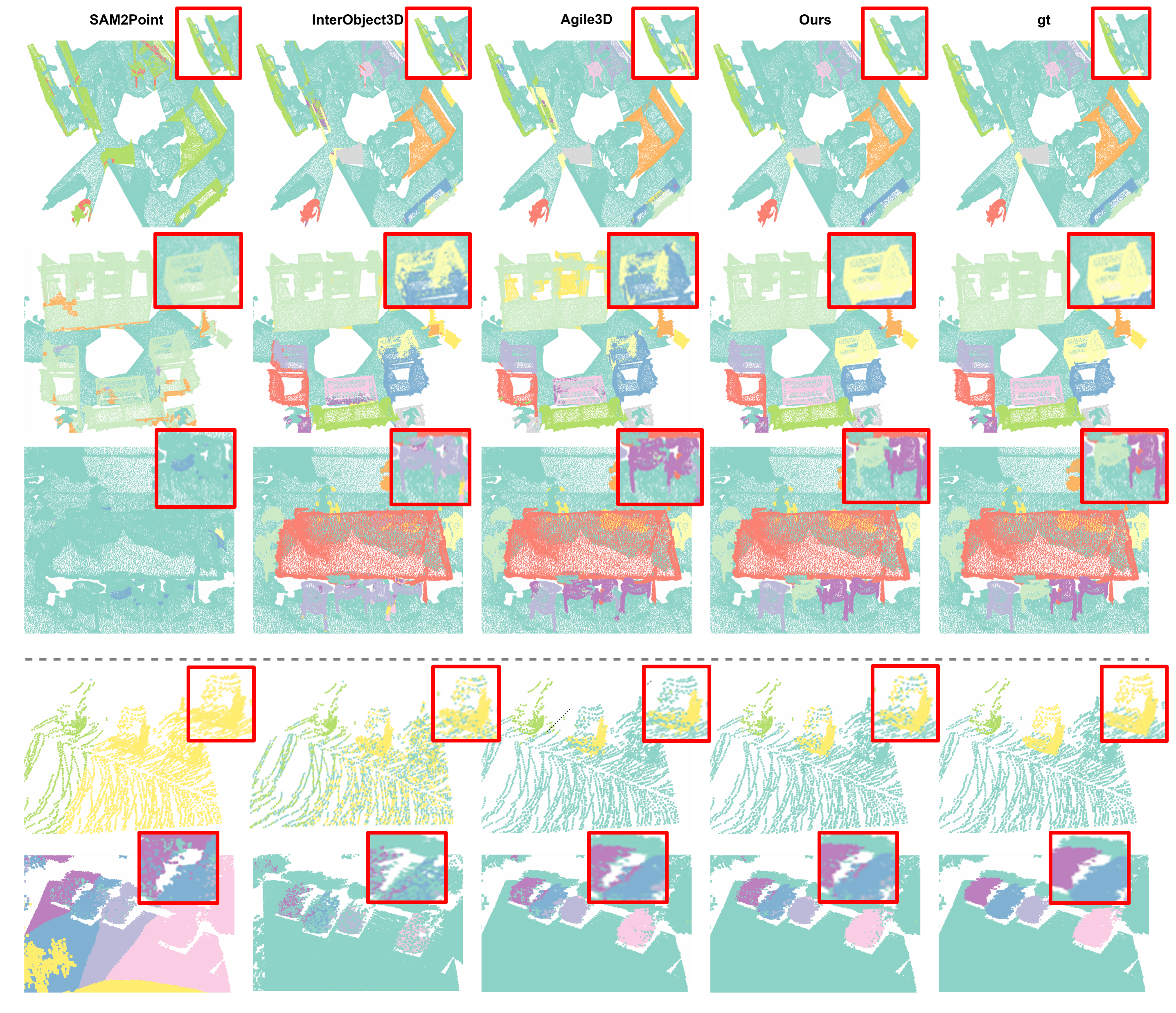}
    \caption{Baseline comparison at 1 click per instance with identical click positions: above the dashed line on ScanNet40~\cite{dai2017scannet}, and below on KITTI360~\cite{Liao2022PAMI}. Each instance class is shown using a consistent color, with the red box showing a zoomed-in region for closer inspection of the segmentation mask details.}
    \label{fig:baseline-cmps}
\end{figure*}

For baseline comparison, we provide both qualitative evaluation results in Fig.~\ref{fig:baseline-cmps} and the quantitative evaluation results in Tab.~\ref{tab:baseline-cmps-clicks} and Tab.~\ref{tab:cmps-baselines}.

\textbf{Qualitative Comparisons}: The baseline comparison in Fig.~\ref{fig:baseline-cmps} demonstrates clear visual improvements across different methods. Our approach produces more accurate and coherent instance segmentation results compared to existing methods. The red boxes highlight specific regions where our method shows superior boundary precision and reduced fragmentation. Notably, InterObject3D struggles with over-segmentation and noise, while SAM2Point and Agile3D show inconsistent instance boundaries. Our method maintains better spatial coherence and produces cleaner segmentation masks across both indoor (ScanNet40) and outdoor (KITTI360) scenarios.

\textbf{Quantitative Comparisons}: Tab.~\ref{tab:baseline-cmps-clicks} reveals that performance consistently improves with more clicks across all methods, although gains diminish after 5-7 clicks. Our method achieves 71.6\% mIoU on ScanNet40 and 64.7\% on KITTI360 with single clicks, significantly outperforming baselines. SAM2Point performs competitively on ScanNet40 (65.2\%) but shows substantial degradation on KITTI360, indicating poor generalization to outdoor scenes. InterObject3D methods perform poorly overall, especially on KITTI360, where they achieve only 2.6-3.4\% mIoU. Tab.~\ref{tab:cmps-baselines} and provides comprehensive metrics showing our method's superiority across all evaluation criteria. The consistent ranking across metrics (Ours > SAM2Point > Agile3D > InterObject3D variants) demonstrates the robustness of our architectural improvements and training strategy.

For additional qualitative and quantitative results evaluated on the two selected datasets, please refer to the supplementary material.

\begin{table}[!th]
\vspace{-1.0em}
    \caption{Metric evaluation results (mIoU ) for object instance segmentation as a function of the number of clicks: top, trained and tested on ScanNet40~\cite{dai2017scannet}; bottom, trained and tested on KITTI360~\cite{Liao2022PAMI}.}
    \label{tab:baseline-cmps-clicks}
\vspace{-0.4em}
\centering
\begin{adjustbox}{width=0.90\linewidth}   
     \begin{tabular}{l l|ccccc}
        \toprule
       Dataset & Method &
        IoU@1$\uparrow$ & IoU@3$\uparrow$ & IoU@5$\uparrow$ & IoU@7$\uparrow$ & IoU@10$\uparrow$ \\
        \midrule
        \multirow{4}{*}{ScanNet40 \cite{dai2017scannet}}
            & InterObj3D & 35.8  & 59.0 & 75.1 & 78.6 & 80.3 \\ 
            & InterObj3D++ & 40.8  & 63.9 & 79.2 & 81.4 & 82.6 \\ 
            & Agile3D & 63.3  & 75.4 & 82.3 & 83.7 & 85.0 \\ 
            & Point-SAM & 52.4 & 76.3 & 81.3 & 84.0 & \textbf{85.7}\\
            & SAM2Point & 65.2 & 73.6 & 78.5 & 80.6 & 84.0 \\ 
            & Ours & \textbf{71.6} & \textbf{79.1} & \textbf{82.5} & \textbf{83.9} & 85.4 \\ 
        \midrule
        \multirow{4}{*}{KITTI360 \cite{Liao2022PAMI}}
            & InterObj3D & 2.6 & 9.8 & 16.4 & 28.5 & 36.3 \\ 
            & InterObj3D++ & 3.4 & 11.0 & 19.9 & 32.6 & 40.5 \\ 
            & Agile3D & 34.8  & 42.7 & 44.4 & 45.8 & 49.6 \\ 
            & Point-SAM & 52.8 & 71.4 & 81.3 & 83.9 & 85.7\\
            & SAM2Point & 49.4  & 74.6 & 81.7 & 84.3 & 85.8 \\ 
            & Ours & \textbf{64.7}  & \textbf{76.2} & \textbf{82.5} & \textbf{85.0} & \textbf{86.1} \\ 
        \bottomrule
    \end{tabular}
\end{adjustbox}
\vspace{-1.4em}
\end{table}

\begin{table}[!th]
\vspace{-0.0em}
\centering
    \caption{Evaluation of instance segmentation using mIoU, mAP@0.25, and mAP@0.5 on ScanNet40~\cite{dai2017scannet} (left) and KITTI360~\cite{2017arXiv170201105A}, showing full metric results.}
    \label{tab:cmps-baselines}
\large
\vspace{-0.4em}
\begin{adjustbox}{width=\linewidth}   
     \begin{tabular}{c | cccc cccc}
        \toprule
        \multirow{2}{*}{Method} & \multicolumn{3}{c}{Scannet40 \cite{dai2017scannet}} & \multicolumn{3}{c}{KITTI360 \cite{Liao2022PAMI}} \\
        \cmidrule(lr){2-4}\cmidrule(lr){5-7}
        & mIOU $\uparrow$ & mAP$_{25\%}$ $\uparrow$ & mAP$_{50\%}$ $\uparrow$ & mIOU $\uparrow$ & mAP$_{25\%}$ $\uparrow$  & mAP$_{50\%}$ \\
        \hline
        InterObject3D & 35.8 & 42.6 & 28.1 & 2.6 & 6.8 & 2.1 \\ 

        InterObject3D++ \cite{kontogianni2022interObj3d}  & 40.8 & 48.2 & 32.1 & 3.4 & 8.5 & 2.8 \\ 

        Agile3D \cite{yue2023agile3d} & 63.3 & 59.8 & 53.5 & 34.8 & 38.4 & 29.2 \\ 

        Point-SAM & 52.4 & 61.7 & 44.3 & 52.8 & 57.1 & 47.4 \\

        SAM2Point \cite{guo2024sam2point} & 65.2 & 62.5 & 55.1 & 49.4 & 54.2 & 41.8 \\

        Ours & $\mathbf{71.6}$ & $\mathbf{68.4}$ & $\mathbf{60.1}$ & $\mathbf{64.7}$ & $\mathbf{69.2}$ & $\mathbf{55.6}$ \\ 
        \bottomrule
    \end{tabular}
\end{adjustbox}    
\vspace{-1.2em}
\end{table}

The cross-dataset evaluation in \cref{tab:multi_object_cross} further shows three clear trends in terms of IoU and NoC. Under the in-domain setting (ScanNet40 $\rightarrow$ ScanNet40), all methods perform strongly, with our model consistently ranking among the best. When transferring to unseen indoor data (ScanNet40 $\rightarrow$ S3DIS), performance drops moderately but remains relatively stable, and our approach maintains a clear advantage in both segmentation accuracy and interaction efficiency. However, in the cross-domain indoor-to-outdoor setting (ScanNet40 $\rightarrow$ KITTI360), all methods suffer a substantial degradation, indicating a significant domain gap, though our model still demonstrates comparatively better robustness.
\begin{table*}[t]
\centering
\caption{Quantitative results on interactive multi-object segmentation. For a fair comparison, some interactive single-object segmentation approaches are adapted to our multi-object evaluation setting. These baseline methods continue to generate binary masks for individual objects, and the resulting masks are manually combined to obtain the final multi-object predictions, which follow a similar processing strategy to that proposed in Agile3D~\cite{yue2023agile3d}.}
\label{tab:multi_object_cross}
\begin{adjustbox}{width=1.0\linewidth}   
\begin{tabular}{l c ccc ccc}
\toprule
\textbf{Method} & \textbf{Train $\rightarrow$ Eval} 
& \textbf{IoU@5$\uparrow$} & \textbf{IoU@10$\uparrow$} & \textbf{IoU@15$\uparrow$}
& \textbf{NoC@80$\downarrow$} & \textbf{NoC@85$\downarrow$} & \textbf{NoC@90$\downarrow$} \\
\midrule
InterObject3D   & \multirow{6}{*}{ScanNet40 $\rightarrow$ ScanNet40} 
& 75.1 & 80.3 & 81.6 & 10.2 & 13.5 & 16.6 \\
InterObject3D++ &  
& 79.2 & 82.6 & 83.3 & 8.6  & 12.4 & 15.7 \\
AGILE3D  &  
& 82.3 & 85.0 & 86.0 & 6.3 & 10.0 & 14.4 \\
Point-SAM   & 
& 81.3 & \textbf{85.7} & 85.9 & 7.1 & 11.2 & 15.0 \\
SAM2Point  &  
& 78.5 & 84.0 & 84.3 & 7.8 & 11.8 & 15.6 \\
Ours       &  
& \textbf{82.5} & 85.4 & \textbf{86.2} & \textbf{6.0} & \textbf{9.6} & \textbf{13.8} \\

\midrule
InterObject3D   & \multirow{6}{*}{ScanNet40 $\rightarrow$ S3DIS-A5} 
& 76.9 & 85.0 & 87.3 & 6.8 & 8.8 & 13.5 \\
InterObject3D++ &  
& 81.9 & 88.3 & 89.3 & 5.7 & 7.6 & 11.6 \\
AGILE3D  &  
& 86.3 & 88.3 & \textbf{90.3} & 3.4 & 5.7 & \textbf{9.6} \\
Point-SAM   &  
& 78.2 & 81.1 & 83.0 & 6.6 & 9.3 & 13.9 \\
SAM2Point  &  
& 75.6 & 78.4 & 81.2 & 7.2 & 10.4 & 14.8 \\
Ours       &  
& \textbf{86.9} & \textbf{88.7} & 90.1 & \textbf{3.2} & \textbf{5.4} & 10.1 \\
\midrule
InterObject3D   & \multirow{6}{*}{ScanNet40 $\rightarrow$ KITTI360} 
& 10.5 & 22.1 & 31.0 & 19.8 & 19.8 & 19.9 \\
InterObject3D++ &  
& 16.7 & 37.1 & \textbf{52.2} & 18.3 & 18.9 & 19.3 \\
AGILE3D  &  
& 40.5 & 44.3 & 48.2 & 17.4 & 18.3 & 18.8 \\
Point-SAM   &
& 24.6 & 30.2 & 35.4 & 18.9 & 19.1 & 19.4 \\
SAM2Point  &  
& 26.1 & 33.7 & 37.9 & 18.4 & 18.8 & 19.2 \\
Ours       &  
& \textbf{41.7} & \textbf{45.6} & 50.8 & \textbf{17.1} & \textbf{17.5} & \textbf{18.0} \\
\bottomrule
\end{tabular}
\end{adjustbox}
\end{table*}

\subsection{Ablation Study}
The ablation study in Tab.~\ref{tab:ablation} highlights the importance of each model component. The choice of encoder strongly affects performance, with the Point V3 Encoder (line 1) outperforming the Sparse U-Net (line 2), demonstrating superior feature extraction. Both semantic embeddings (line 3) and spatial embeddings (line 4) contribute substantially to performance. The mask head is critical for accurate spatial prediction (line 6), while the class head primarily improves semantic consistency (line 5). Loss function ablations show CE Loss (line 10) is most important, whereas BCE Loss (line 8) and Dice Loss (line 9) have a moderate influence. The Query Adaptor (line 7) provides refinement with a minor drop when removed. Sampling strategy also matters: Farthest Point Sampling (line 13) preserves performance, while Random and Voxel Sampling degrade results.

\begin{figure}[!th]
\vspace{-0.0em}
    \centering
    \includegraphics[width=\linewidth]{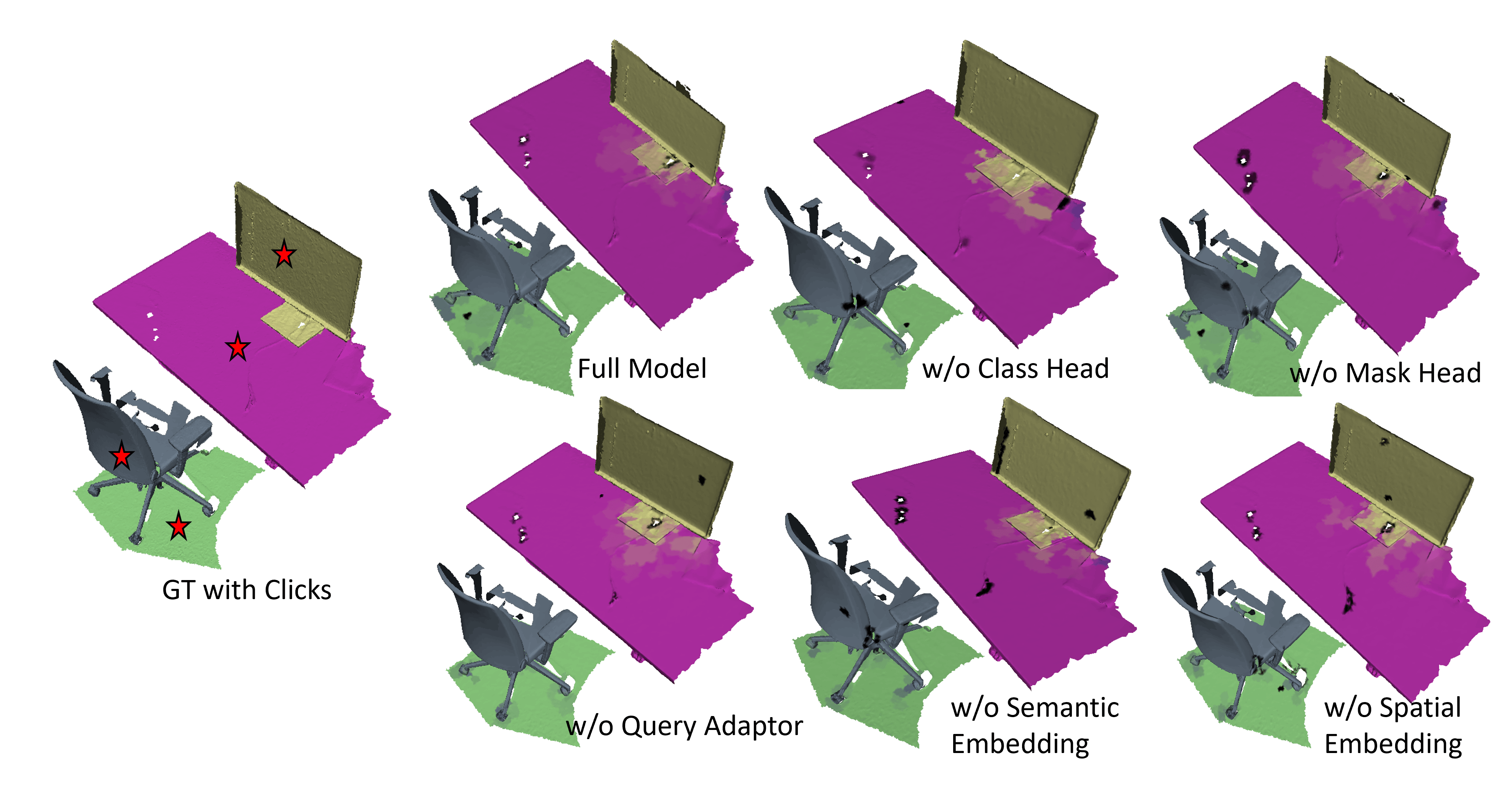}
    \caption{Ablation study visualization of instance segmentation on a selected indoor scene with different modules removed; the leftmost shows the Ground Truth, with stars indicating click point positions as input.}
    \label{fig: viz-ablation of main module}
    \vspace{-0.6em}
\end{figure}

We visualize lines 3--7 in Fig.~\ref{fig: viz-ablation of main module}. Removing the class head produces minor misclassifications, whereas mask head removal leads to inconsistent segmentation and missing regions. Spatial embedding removal causes boundary imprecision and instance merging, while semantic embedding removal results in inconsistent class assignments. Query Adaptor removal yields marginal boundary degradation. These visual patterns align with Tab.~\ref{tab:ablation}, validating the necessity of each component for robust segmentation.

Tab.~\ref{tab:model-complexity} shows our method balances efficiency and performance. While some methods offer faster inference or smaller model size, our approach achieves competitive speed and reasonable memory footprint, enabling real-time, high-quality segmentation without iterative processing.

\begin{table}[!th]
\vspace{-0.2em}
\centering
    \caption{Ablation study on the main modules of our model, evaluated on the ScanNet40 \cite{dai2017scannet} test split.}
    \label{tab:ablation}
\begin{adjustbox}{width=0.76\linewidth}    
     \begin{tabular}{l c c c c}
        \toprule 
       & & mIOU $\uparrow$ & mACC $\uparrow$ & mAP$_{50\%}$ $\uparrow$ \\
        \hline
        \textbf{1.} & Point V3 Encoder & $\mathbf{71.6}$ & $\mathbf{78.2}$ & $\mathbf{60.1}$  \\
        \textbf{2.} & Sparse U-Net Encoder & 45.8 & 55.2 & 38.4 \\
        \hline
        \textbf{3.} & w/o Semantic Embedding & 52.3 & 62.1 & 43.8 \\
        \textbf{4.} & w/o Spatial Embedding & 48.7 & 58.6 & 40.2 \\
        \textbf{5.} & w/o Class Head & 41.2 & 51.8 & 33.7 \\
        \textbf{6.} & w/o Mask Head & 36.8 & 46.9 & 29.4 \\
        \textbf{7.} & w/o Query Adaptor & 67.2 & 74.8 & 56.3 \\
        \hline
        \textbf{8.} & w/o BCE Loss & 64.8 & 72.1 & 54.2 \\
        \textbf{9.} & w/o Dice Loss & 68.3 & 75.6 & 57.8 \\
        \textbf{10.} & w/o CE Loss & 53.4 & 63.2 & 44.6 \\
        \hline
        \textbf{11.} & Random Sampling & 48.0 & 57.4 & 40.5 \\
        \textbf{12.} & Voxel Sampling & 66.4 & 73.2 & 55.7 \\
        \textbf{13.} & Farthest Point Sampling & $\mathbf{71.6}$ & $\mathbf{78.2}$ & $\mathbf{60.1}$ \\
        \bottomrule
    \end{tabular}
\end{adjustbox}
\vspace{-0.0em}
\end{table}

\begin{table}[!thbp]
\vspace{-0.0em}
\centering
\caption{Comparison of model complexity and inference time with key baselines.}
\label{tab:model-complexity}
    \vspace{-0.6em}
\begin{small}
\begin{adjustbox}{width=0.7\linewidth}
\begin{tabular}{c c c c c }
\hline
 & Agile3D & Point-SAM & InterObject3D & Ours\\%
\hline
Model Size (Mb) $\downarrow$ & 151.3 & 1240.2 & 144.6 & $\mathbf{128.5}$ \\
Latency(s) $\downarrow$ & 1.82 & 0.41 & 0.08 & $\mathbf{0.06}$\\ 
\hline
\end{tabular}
\end{adjustbox}
\end{small}
\end{table}

\begin{figure}[!htbp]
    \centering
    \vspace{-1.5em}
    \includegraphics[width=0.94\linewidth]{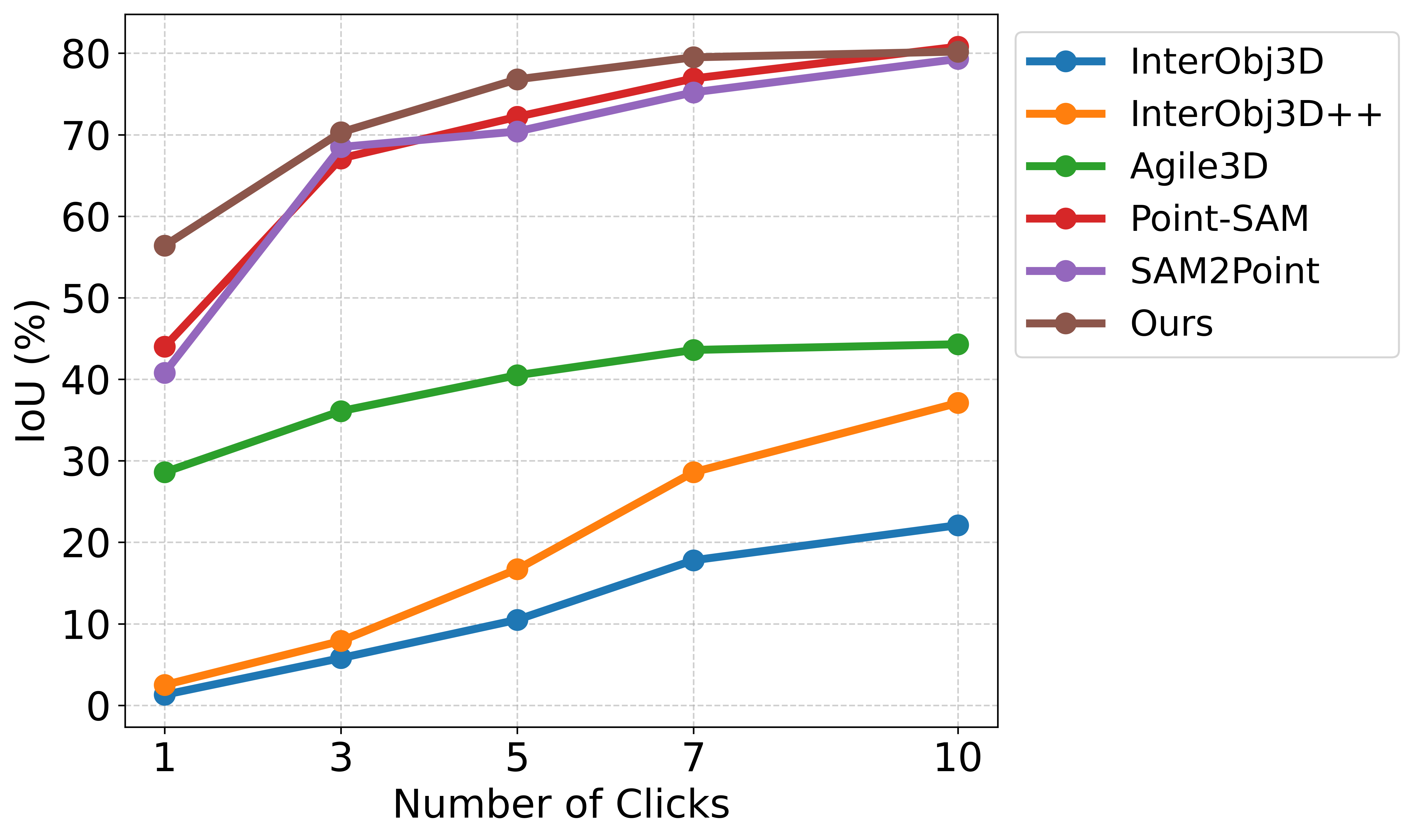}
    \vspace{-1.0em}
    \caption{Plot of mIoU results for all methods as a function of the number of clicks.}
    \label{fig: cross eval of clicks}
    \vspace{-2.0em}
\end{figure}

Fig.~\ref{fig: cross eval of clicks} compares baseline methods as the number of clicks increases from 1 to 10. Our model consistently achieves the best performance, reaching about 80\% mIoU, while Point-SAM and SAM2Point follow. InterObj3D and Agile3D show smaller gains and plateau earlier. Across all methods, performance improves sharply within the first 3–4 clicks before gradually leveling off, with our approach showing the largest margin in the few-click setting (less than 5 clicks).

\begin{figure}[!th]
    \centering
    \includegraphics[width=0.66\linewidth]{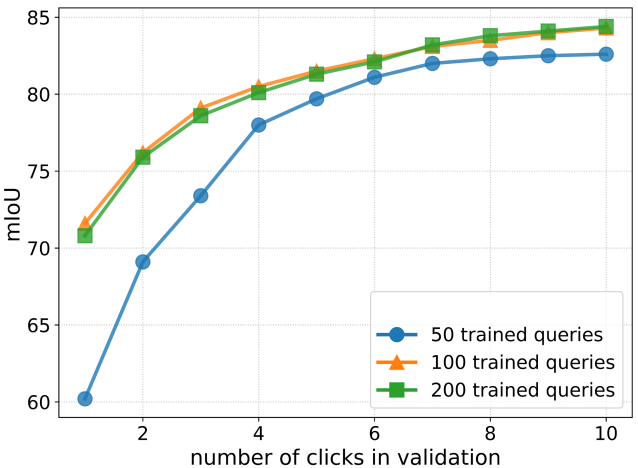}
    \vspace{-0.8em}
    \caption{Plot of mIoU test performance on ScanNet40 as a function of the number of click query points during inference (The query numbers ranging from 50 to 200 during training are explored).}
    \label{fig: ablation of training click number}
    \vspace{-2.0em}
\end{figure}

Fig.~\ref{fig: ablation of training click number} shows performance scaling with different training data volumes (50, 150, and 250 queries). Larger training sets consistently yield better results, with the 250-query model reaching about 83\% mIoU versus ~80\% for smaller sets. All configurations follow similar curves, improving rapidly in the first few clicks and saturating around 7–10 clicks, indicating that both data diversity and user feedback enhance segmentation accuracy.

\begin{figure}[!th]
    \centering
    \includegraphics[width=\linewidth]{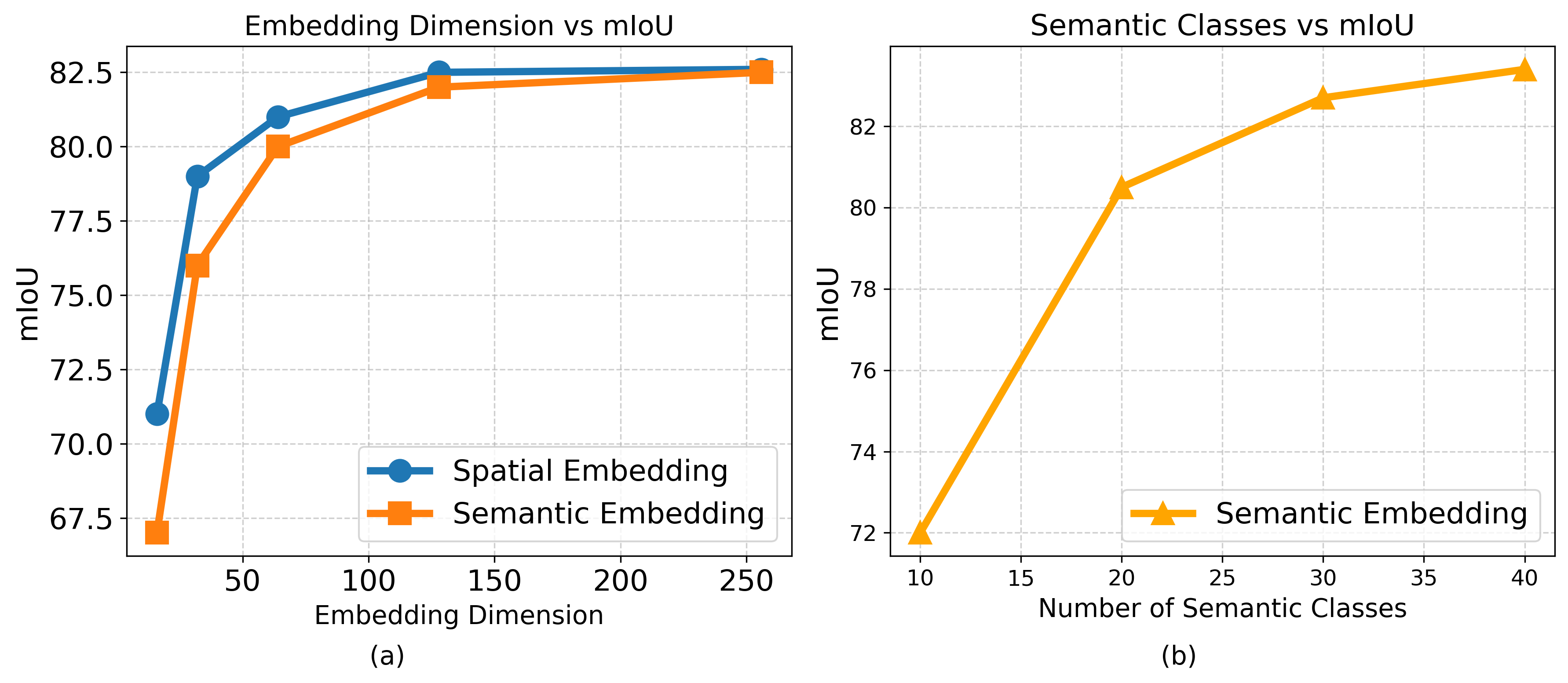}
    \vspace{-1.2em}
    \caption{Plot of (a) mIoU as a function of embedding dimension, with spatial embeddings in blue and semantic embeddings in orange; (b) mIoU as a function of the number of semantic class prototype embedding.}
    \label{fig: ablation of embedding dim-num}
    \vspace{-2.0em}
\end{figure}

Fig.~\ref{fig: ablation of embedding dim-num} analyzes embedding dimensionality and semantic prototype number. Performance rises quickly with dimension size, saturating around 256–512 dimensions at ~82.5\% mIoU. The model also maintains robust accuracy as the number of semantic classes grows, with clear saturation trends showing that moderate embedding dimensions capture most semantic feature information while supporting larger semantic information representation.

\section{Conclusion}
We presented a single-forward-pass interactive 3D segmentation framework that unifies click-guided query learning with semantic prototyped-conditioned refinement. By eliminating iterative re-inference and explicitly modeling inter-instance competition, our method significantly improves few-click efficiency and cross-dataset generalization. 

\textbf{Limitations \& future work} Despite its strong performance, the current framework is limited to point-cloud–based click queries and does not yet exploit richer cross-modal cues, such as text prompts or SAM-derived 2D mask guidance. In addition, user interaction is still required to initiate segmentation, and the model does not explicitly reason about fine-grained part hierarchies. Future work will extend the framework to multi-modal interactive segmentation, integrating text, image, and 2D foundation-model prompts to further reduce annotation effort. 

%
%
\bibliographystyle{splncs04}
\bibliography{main}

\vspace{1em}

\clearpage
\setcounter{section}{0}
\setcounter{figure}{0}
\setcounter{table}{0}
\setcounter{equation}{0}

\renewcommand{\thesection}{S\arabic{section}}
\renewcommand{\thefigure}{S\arabic{figure}}
\renewcommand{\thetable}{S\arabic{table}}
\renewcommand{\theequation}{S\arabic{equation}}

\begin{center}
    \LARGE\bfseries Supplementary Material: Few-Click-Driven Interactive 3D Segmentation with Semantic Embedding
\end{center}
\section{Experiments \& Results}
\begin{figure*}[!th]
    \centering
\includegraphics[width=\linewidth]{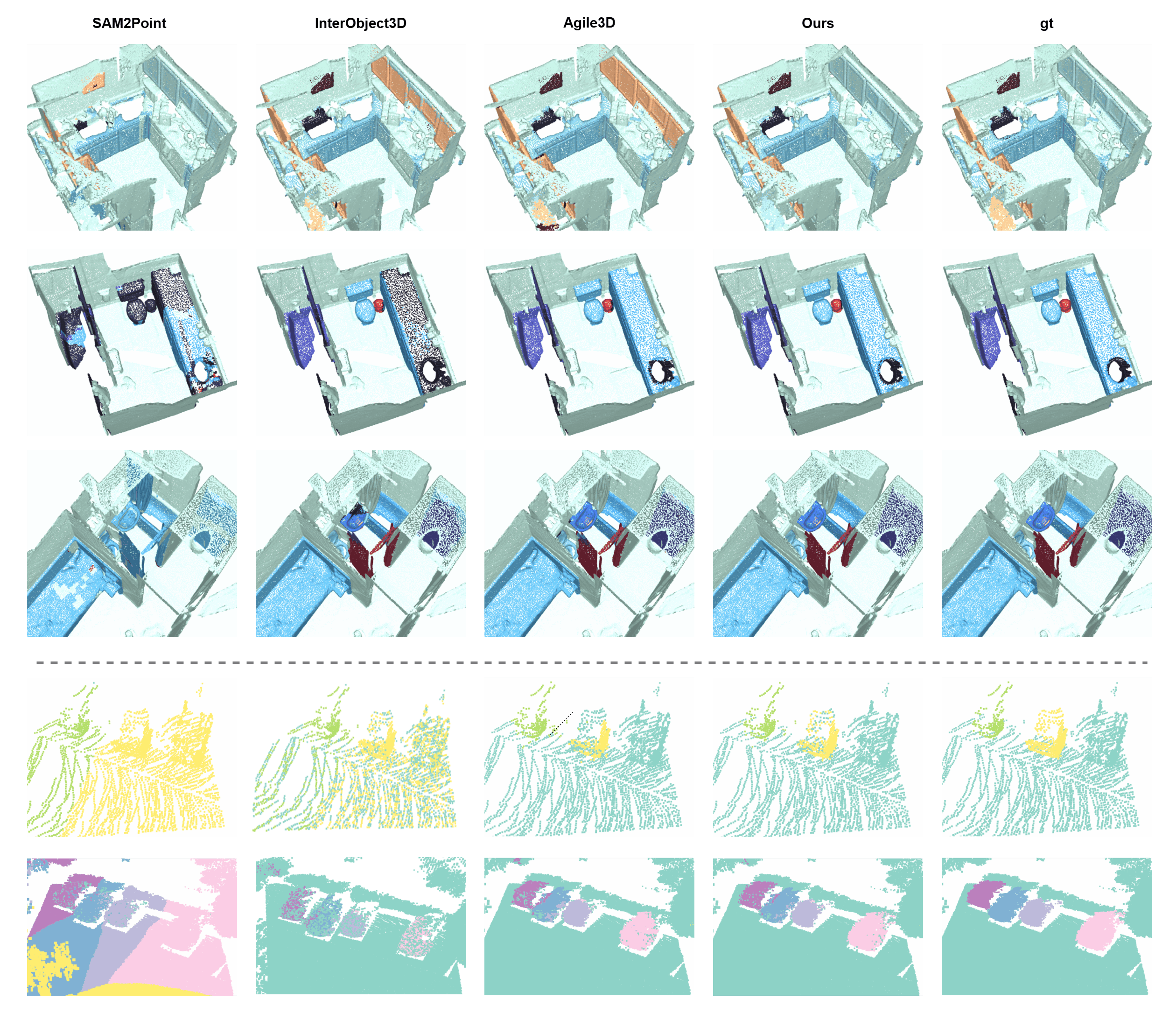}
    \caption{Baseline comparison at 1 click per instance with identical click positions: above the dashed line on ScanNet40~\cite{dai2017scannet}, and below on KITTI360~\cite{Liao2022PAMI}. Each instance class is shown using a consistent color.}
    \label{fig:baseline-cmps-supplem}
\end{figure*}

Qualitative comparison of interactive 3D segmentation results is presented in \cref{fig:baseline-cmps-supplem}. We compare SAM2Point~\cite{guo2024sam2point}, InterObject3D~\cite{kontogianni2022interObj3d}, Agile3D~\cite{yue2023agile3d}, and our method against ground truth (gt). The upper dashed region shows results on ScanNet40~\cite{dai2017scannet} and KITTI360~\cite{Liao2022PAMI}, while the lower part presents additional challenging scenes. Our method produces more accurate instance boundaries and cleaner multi-object separation, closely matching ground truth across diverse indoor and outdoor scenes.

Table~\ref{tab:single_object_extended} summarizes the quantitative results on binary single-object interactive segmentation across in-domain and cross-dataset settings. Overall, our method consistently achieves the best or highly competitive IoU while requiring fewer clicks than all baselines. In the in-domain setting (ScanNet40 $\rightarrow$ ScanNet40), our approach delivers the highest segmentation accuracy while maintaining lower interaction cost. Under the indoor cross-dataset evaluation (ScanNet40 $\rightarrow$ S3DIS-A5), performance drops slightly for all methods, yet our model retains a clear advantage in both IoU and NoC, demonstrating strong generalization in indoor environments. In the more challenging cross-domain setting (ScanNet40 $\rightarrow$ KITTI-360), all methods suffer a substantial performance degradation due to the indoor–outdoor domain gap; nevertheless, our approach remains the most robust, achieving superior accuracy and interaction efficiency among all compared methods.
\begin{table*}[t]
\centering
\caption{Extended quantitative results on interactive single-object segmentation with baseline methods.}
\label{tab:single_object_extended}
\begin{adjustbox}{width=\linewidth}
\begin{tabular}{l l c c c c c c}
\toprule
\textbf{Method} & \textbf{Train $\rightarrow$ Eval} 
& \textbf{IoU@5$\uparrow$} & \textbf{IoU@10$\uparrow$} & \textbf{IoU@15$\uparrow$} 
& \textbf{NoC@80$\downarrow$} & \textbf{NoC@85$\downarrow$} & \textbf{NoC@90$\downarrow$} \\
\midrule

\multicolumn{8}{c}{\textit{ScanNet40 $\rightarrow$ ScanNet40}} \\

InterObject3D     &  & 72.4 & 79.9 & 82.4 & 8.9 & 11.2 & 14.2 \\
InterObject3D++   &  & 78.0 & 82.9 & 84.2 & 7.7 & 10.0 & 13.2 \\
AGILE3D           &  & 79.9 & 83.7 & 85.0 & 7.1 & 9.6 & 12.9 \\
Point-SAM         &  & 79.0 & 84.5 & 85.2 & 7.4 & 10.6 & 14.0 \\
SAM2Point         &  & 76.8 & 82.8 & 84.0 & 7.9 & 11.0 & 14.6 \\
\textbf{Ours}     &  & \textbf{80.2} & \textbf{84.2} & \textbf{85.5} & \textbf{6.8} & \textbf{9.2} & \textbf{12.5} \\

\midrule

\multicolumn{8}{c}{\textit{ScanNet40 $\rightarrow$ S3DIS-A5}} \\

InterObject3D     &  & 72.4 & 83.6 & 88.3 & 6.8 & 8.4 & 11.0 \\
InterObject3D++   &  & 80.8 & 89.2 & 91.5 & 5.2 & 6.7 & 9.3 \\
AGILE3D           &  & 83.5 & 88.2 & 89.5 & 4.8 & 6.4 & 9.5 \\
Point-SAM         &  & 75.5 & 80.5 & 83.5 & 6.4 & 8.8 & 12.8 \\
SAM2Point         &  & 73.0 & 78.0 & 81.5 & 6.9 & 9.7 & 13.5 \\
\textbf{Ours}     &  & \textbf{84.0} & \textbf{89.0} & \textbf{90.0} & \textbf{4.5} & \textbf{6.0} & \textbf{9.2} \\

\midrule

\multicolumn{8}{c}{\textit{ScanNet40 $\rightarrow$ KITTI-360}} \\

InterObject3D     &  & 14.3 & 26.3 & 35.0 & 19.1 & 19.4 & 19.7 \\
InterObject3D++   &  & 19.9 & 40.6 & 55.1 & 17.0 & 17.7 & 18.4 \\
AGILE3D           &  & 42.0 & 46.0 & 50.0 & 16.5 & 17.8 & 18.3 \\
Point-SAM         &  & 23.5 & 29.5 & 34.0 & 18.5 & 18.9 & 19.3 \\
SAM2Point         &  & 25.0 & 32.0 & 37.0 & 18.0 & 18.5 & 19.0 \\
\textbf{Ours}     &  & \textbf{43.5} & \textbf{47.5} & \textbf{51.5} & \textbf{16.0} & \textbf{17.2} & \textbf{17.8} \\

\bottomrule
\end{tabular}
\end{adjustbox}
\end{table*}

\begin{figure*}[!th]
    \centering
\includegraphics[width=\linewidth]{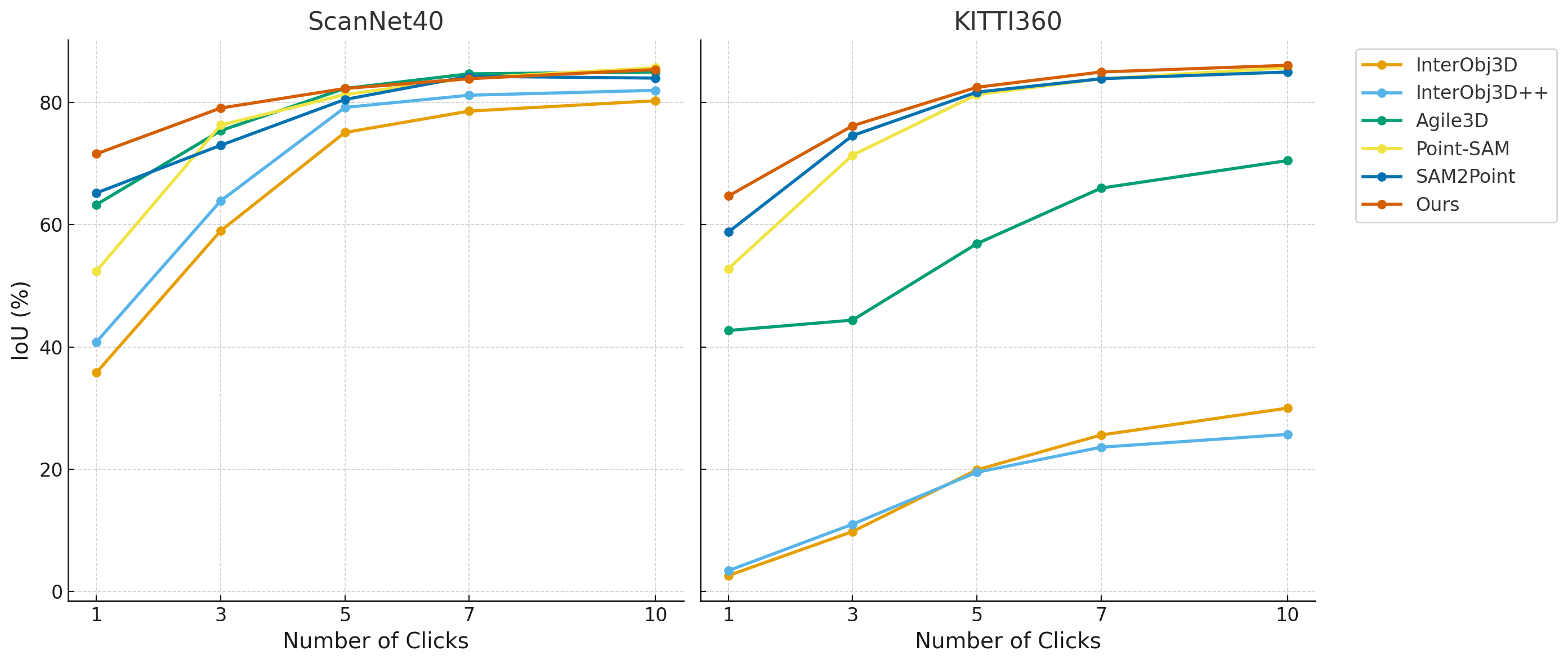}
    \caption{Object instance segmentation IoU (\%) versus number of user clicks on ScanNet40 (left) and KITTI-360 (right). The results compare interaction efficiency across methods, showing how segmentation accuracy improves with additional clicks. Our method achieves higher IoU with fewer interactions and converges faster on both datasets.}
    \label{fig:click-num-supplem}
\end{figure*}

Ablation results on user interaction efficiency are provided in \cref{fig:click-num-supplem}. The dashed upper part reports object instance segmentation IoU (\%) as a function of the number of clicks on ScanNet40 and KITTI-360. Our method consistently achieves higher IoU with fewer clicks and saturates faster than baselines on both datasets, demonstrating more effective utilization of sparse user interactions for accurate multi-object segmentation.

\begin{figure*}[!th]
    \centering
\includegraphics[width=0.90\linewidth]{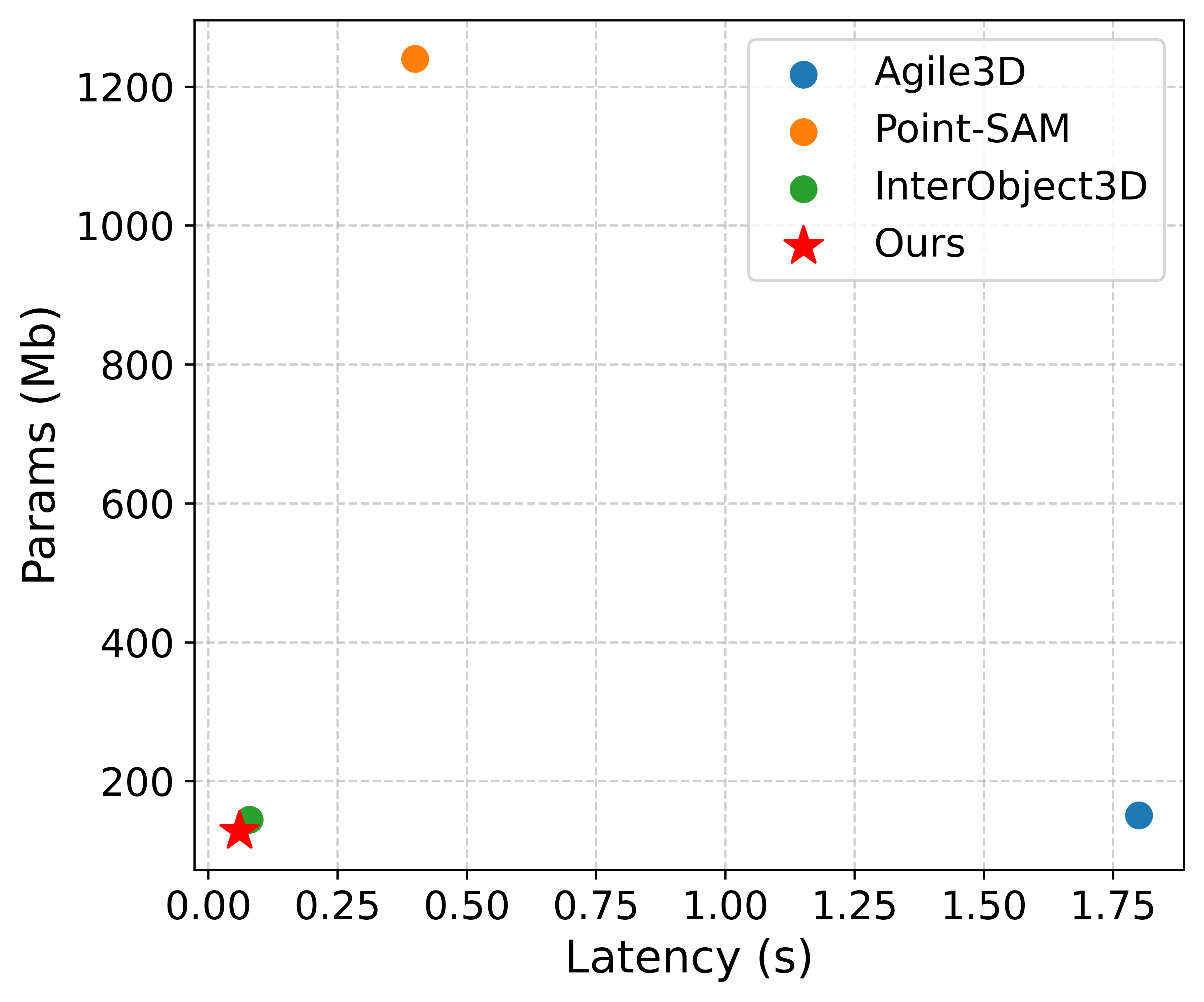}
    \caption{Model efficiency comparison. Memory consumption (parameter size in MB) versus average inference latency (seconds) for different interactive 3D segmentation methods. Our model achieves the most favorable trade-off, with low memory usage and fast inference.}
    \label{fig:time-complexity}
\end{figure*}

The plot in \cref{fig:time-complexity} compares model complexity and efficiency across methods in terms of parameter memory (MB) and average inference latency (s). Point-SAM exhibits the highest memory footprint, while Agile3D incurs the largest latency. InterObject3D reduces latency but remains moderately sized. In contrast, our model achieves the best trade-off, combining low memory consumption with the fastest inference, demonstrating superior efficiency for practical interactive 3D segmentation.

\end{document}